\documentclass[10pt,twocolumn,letterpaper]{article}

\usepackage{iccv}
\usepackage{times}
\usepackage{epsfig}
\usepackage{graphicx}
\usepackage{amsmath}
\usepackage{amssymb}
\usepackage{float}
\usepackage{booktabs}
\usepackage{array}
\usepackage{cuted}\stripsep -3pt plus 3pt minus 2pt
\usepackage{caption}

\usepackage[breaklinks=true,bookmarks=false]{hyperref}

\iccvfinalcopy % *** Uncomment this line for the final submission

\def\iccvPaperID{****} % *** Enter the ICCV Paper ID here

\ificcvfinal\fi

\begin{document}

%%%%%%%%% TITLE
\title{An Efficient Network Design for Face Video Super-resolution}

\author{Feng Yu\\
Joint International Research Laboratory\\ of Information Display and Visualization,\\ Southeast University\\
Nanjing, CN\\
{\tt\small 220181339@seu.edu.cn}
\and
He Li\thanks{Corresponding author}\\
Department of Engineering,\\ University of Cambridge,\\ Cambridge, UK\\
{\tt\small he.li@ieee.org}
\and 
Sige Bian\\
Joint International Research Laboratory\\ of Information Display and Visualization,\\ Southeast University\\
Nanjing, CN\\
{\tt\small 213173744@seu.edu.cn}
\and
Yongming Tang\\
Joint International Research Laboratory\\ of Information Display and Visualization,\\ Southeast University\\
Nanjing, CN\\
{\tt\small tym@seu.edu.cn}
}

\maketitle
\ificcvfinal\thispagestyle{empty}\fi

%%%%%%%%% ABSTRACT
\begin{abstract}
   Face video super-resolution algorithm aims to reconstruct realistic face details through continuous input video sequences. However, existing video processing algorithms usually contain redundant parameters to guarantee different super-resolution scenes. In this work, we focus on super-resolution of face areas in original video scenes, while rest areas are interpolated. This specific super-resolved task makes it possible to cut redundant parameters in general video super-resolution networks. We construct a dataset consisting entirely of face video sequences for network training and evaluation, and conduct hyper-parameter optimization in our experiments. We use three combined strategies to optimize the network parameters with a simultaneous train-evaluation method to accelerate optimization process.
   Results show that simultaneous train-evaluation method improves the training speed and facilitates the generation of efficient networks. The generated network can reduce at least 52.4\% parameters and 20.7\% FLOPs, achieve better performance on PSNR, SSIM compared with state-of-art video super-resolution algorithms. When processing $36\times 36\times 1\times 3$ input video frame sequences, the efficient network provides 47.62 FPS real-time processing performance. We name our proposal as hyper-parameter optimization for face Video Super-Resolution (HO-FVSR), which is open-sourced at https://github.com/yphone/efficient-network-for-face-VSR.
\end{abstract}

\section{Introduction}
The performance improvement of neural network algorithms has brought disruptive breakthroughs to various applications in daily life. Taking face video super-resolution (VSR) as an example, VSR algorithms based on neural networks can achieve better feature extraction and perceptual effect than traditional algorithms. By utilizing the nonintrusive and natural characteristics~\cite{1} of faces, face super-resolution has a great potential to generate an efficient network design with less parameters and operations compared with general VSR algorithms. However, it is a time-consuming and error prone process for human experts to develop efficient architectures currently. Therefore, interests in automated neural architecture search methods have risen~\cite{2}. In our work, we use automatic hyper-parameter optimization strategies to create efficient and reliable face video super-resolution network for different applications.

Single image super-resolution (SISR) algorithms based on neural networks obtain state-of-the-art performance on visual perception and peak signal-to-noise ratios in~\cite{3,4,5}. In the optimized design of network architecture, the appropriate combinations of basic layers, such as up-sampling, convolution, pooling \emph{etc.}, are particularly important. 
For VSR models, networks composed of optical flow module, residual module and up-sample module~\cite{6,7} have been presented, which usually use a fixed combination of basic layers to extract the effective information among contiguous frames. 
%All sub-modules are composed of different basic layers. 
These fixed neural network architectures can be extended to many super-resolution application fields. 
The model hyper-parameters such as the number of different submodules and convolutional channels in basic layer can be adjusted, therefore, they have great design flexibility. 
In this paper, we use the tree-structured parzen estimator (TPE)~\cite{8}, random search~\cite{9}, sequential model-based algorithm configuration (SMAC)~\cite{10} strategies to optimize the hyper-parameters of VSR neural network architectures for an efficient network design. We create a face dataset for training and evaluation. The dataset includes consecutive face sequences extracted from original videos. However, The optimization process for an efficient network deisgn based on face dataset usually elapses quantity of time. To address this problem, a simultaneous train-evaluation method is applied to accelerate the optimization process. We estimate the performance of candidate hyper-parameters after several training epochs. The evaluation results are used by search strategies to generate next candidate hyper-parameters. Finally, simultaneous train-evaluation method considers the performance evaluation values of all candidate hyperparameters and selects the optimal hyperparameter to construct efficient network.
%In the initial experiment stage, the number of candidate hyper-parameters and training epochs are both set to fixed values to limit the optimization process. Results show that the simultaneous train-evaluation method elapses less time to facilitate the generation of efficient network, the efficient network illustrates excellent reductions in parameters and operations, better performance on face super-resolution tasks.

The contributions of this paper are as follows:
\begin{itemize}
   \item We create a specific face video sequence dataset for network training and evaluation.
   \item We design a simultaneous train-evaluation method for hyper-parameter optimization process to reduce time consumption.
   \item We perform hyper-parameter optimization to generate an efficient network, HO-FVSR, which achieves at least 52.4\% parameters, 20.7\% FLOPs reduction and better performance on PSNR, SSIM versus the state-of-the-art works~\cite{6,7}.
   \item The efficient network provides 47.62 FPS real-time processing performance when processing $36\times 36\times 1\times 3$ input video frame sequences.
\end{itemize}

\section{Related work}
\subsection{Video Super-resolution Network Design}
The video super-resolution algorithms are evolved from the single-image super-resolution algorithms. 
State-of-the-art single image super-resolution methods tried to exploit internal similarities of the same image~\cite{11} to learn mapping functions from low- and high- resolution exemplar pairs~\cite{12}. In 2014, Dong~\emph{et al.} first used convolutional neural networks (CNNs) to replace traditional algorithms to solve super-resolution problems, called SRCNN~\cite{3}. SRCNN~\cite{3} significantly improved performance in terms of peak signal-to- noise ratio (PSNR). After that, many image super-resolution algorithms are proposed by using convolutional neural networks, such as FSRCNN~\cite{4}, EDSR~\cite{5}. In video applications, researchers usually use specific designed convolutional submodules to estimate the motion vector between continuous video frames and fuse multi-frame video frames, which has significant advantages compared with image algorithms. In 2017, Caballero~\emph{et al.} proposed a real-time video-based super-resolution algorithm (VESPCN~\cite{13}). They realized the super-resolution processing of self-contained continuous video images by adding a motion compensation module to the front end of single image network, which is better than SRCNN. In the same year, Tao~\emph{et al.} proposed to use a sub-pixel motion compensation network (SPMC~\cite{14}) to implement a highly versatile video super-resolution algorithm. Recently, Wang~\emph{et al.} proposed a coarse-to-fine optical flow estimation network to extract relevant information between consecutive video frames, and used it in super-resolution process to achieve better video super-resolution results~\cite{6, 7}. These video super-resolution networks can be categorized to the combination of several submodules with different functions, which contains residual submodule for image feature extraction, up-sample submodule for resizing images and optical flow submodule for information extraction among video frames. In this paper, we take the submodule design for reference, use a fixed neural network architecture to optimize the network parameters.

\subsection{Hyper-parameter Optimization}
\begin{figure}[h!]
   \centering
   \includegraphics[width=\columnwidth]{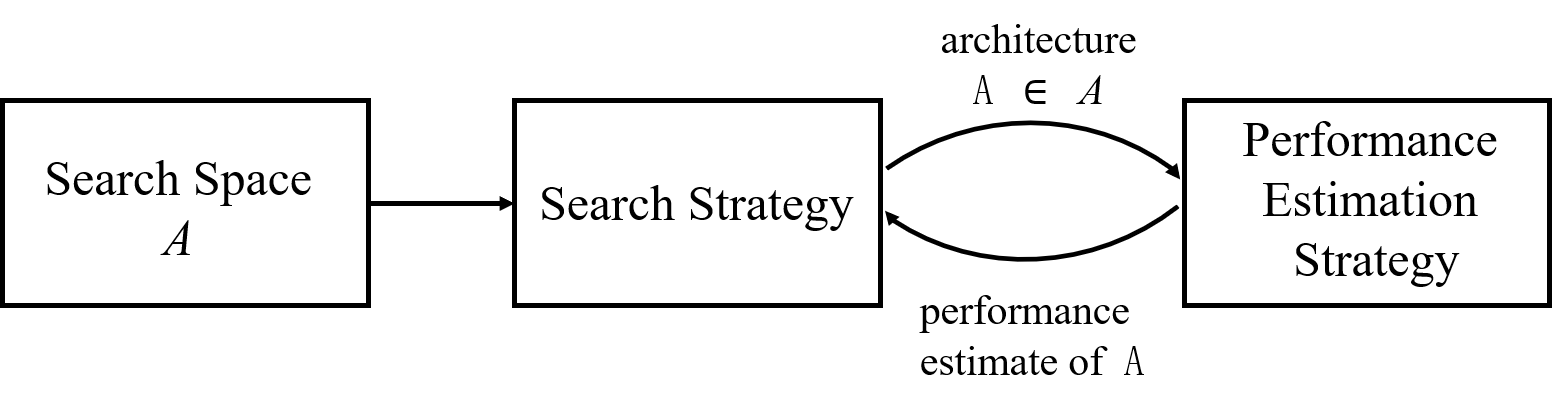}
   \caption{Illustration of neural architecture search methods~\cite{2}. A search strategy selects an architecture from a predefined search space. The architecture is passed to a performance estimation strategy, which returns the estimated performance of selected architecture to the search strategy.}
   \label{fig1}
\end{figure}

Hyper-parameter optimization has significant overlap with neural architecture search, and can be categorized into three dimensions: search space, search strategy and performance estimation~\cite{2}. As shown in Figure \ref{fig1}, the search space defines the network architectures which can be represented in principle; it usually incorporates with prior knowledge about typical properties of architectures well-suited for search space reduction. The search strategy specifies how to explore the search space. It usually works in conjunction with the performance estimation strategy to generate candidate architectures. 
\begin{figure*}[ht]
   \centering
   \includegraphics[width=\textwidth]{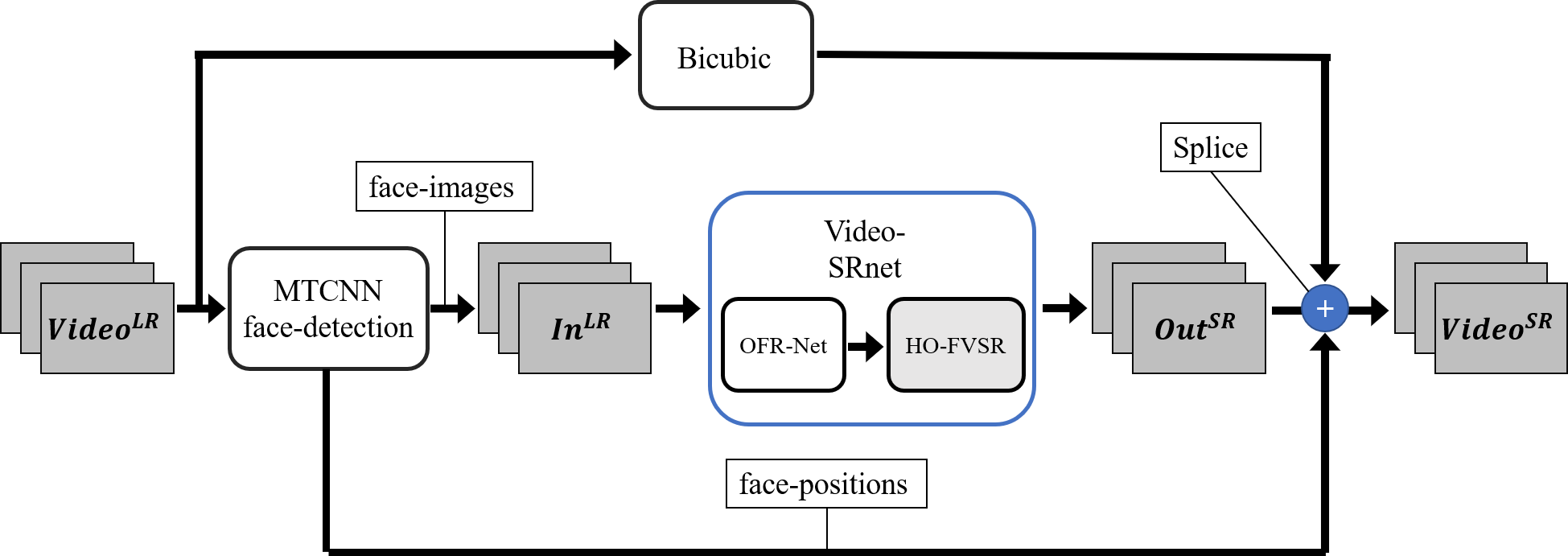}
   \caption{The overall framework of face video super-resolution algorithm. In this framework, the face area needs to be processed by video super-resolution algorithm which consists of OFR-net and HO-FVSR, where OFR-net denotes optical flow network. The rest of the video frames are interpolated using bicubic. The final output image splices the two processed images.}
   \label{fig2}
\end{figure*}

% \begin{strip}
%    \centering\includegraphics[width=\textwidth]{figure2}
%    \captionof{figure}{The overall framework of face video super-resolution algorithm. In this framework, the face area needs to be processed by video super-resolution algorithm which consists of OFR-net and HO-FVSR. OFR-net means optical flow network and HO-FVSR is the abbreviation of hyper-parameter for face video super-resolution. The rest of the video frames are interpolated using bicubic. The final output image splices the two processed images.}
%    \label{fig2}
% \end{strip}

Grid search and manual search are the most widely used strategies for hyper-parameter optimization. Tree-structured parzen estimator approach (TPE)~\cite{8} uses Bayesian rules to build a model, by transforming generative process and replacing the distributions of configuration prior with non-parametric densities to realize the search optimization design of network parameters. Bergstra and Bengio showed empirically and theoretically that randomly chosen trials are more efficient for hyper-parameter optimization than trials on a grid~\cite{9}. Random search method over the same domain is able to find optimised models within a small fraction of the computation time. Hutter~\emph{et al.} proposed sequential model-based algorithm configuration (SMAC)~\cite{10}, which can be understood as an extension of  a racing algorithm that selects configurations based on a model rather than uniformly at random. The SMAC method can be applied to optimize the solution quality within a fixed time budget.

\subsection{Facial Feature Extraction}
The face structure has natural and non-invasive characteristics~\cite{1}, which are mainly reflected in the similarity of shape and the differences of different faces.  Reasonable use of facial features will constraint super-resolution process to a specific domain and is able to decrease network complexity. The current algorithms for facial feature extraction are mainly based on the realization of deep neural networks and computer vision. Representative algorithms are MTCNN~\cite{15} and Dlib libraries\footnote{http://dlib.net/}. MTCNN uses three cascaded neural network structures to estimate the facial features from coarse to fine, and finally outputs the candidate face area. The Dlib library returns 68 key point features of face and candidate face area, however limitations in execution speed and framework compatibility exist. 
Therefore, the proposed HO-FVSR used MTCNN to generate dataset.

\section{Methodology}
The essential step to realize efficient network design is to produce a specific dataset for training and evaluation process. We first split the original video containing clear faces to continuous frame sequences, then use MTCNN~\cite{15} to select faces on each video frame, and resize the obtained target face images to a fixed size. Finally, the fixed-size image sequences are stored in dataset directory, as the face dataset used for training and evaluation.

\subsection{Framework of Face Video Super Resolution Algorithm}
Face video super-resolution algorithm mainly targets the face areas in original videos, while does not consider other areas. Therefore, in the implementation of VSR algorithms, it is necessary to circle and extract the low-resolution face areas in
original video frames, record their positions. The low-resolution face sequences are super-resolved by face VSR algorithms, other areas are only interpolated into a corresponding size. These two partial images are merged as final output. The overall framework of face VSR is shown in Figure \ref{fig2}. We use MTCNN to delineate the face image in the original video and record the coordinate position. At the same time, since VSR algorithm requires a continuous sequence of video frames as input, in this paper, three continuous images are needed as input, \emph{i.e.} for low-resolution video images at time $T$. It is necessary to use the video frames at time $T+1$ and time $T-1$. As a result, in the stage of real-time video processing, the input frame video buffer need to be delayed by one frame to meet the needs of real-time processing.

\subsection{The Design of Efficient Network}
\begin{figure}[ht]
   \centering
   \includegraphics[width=\columnwidth]{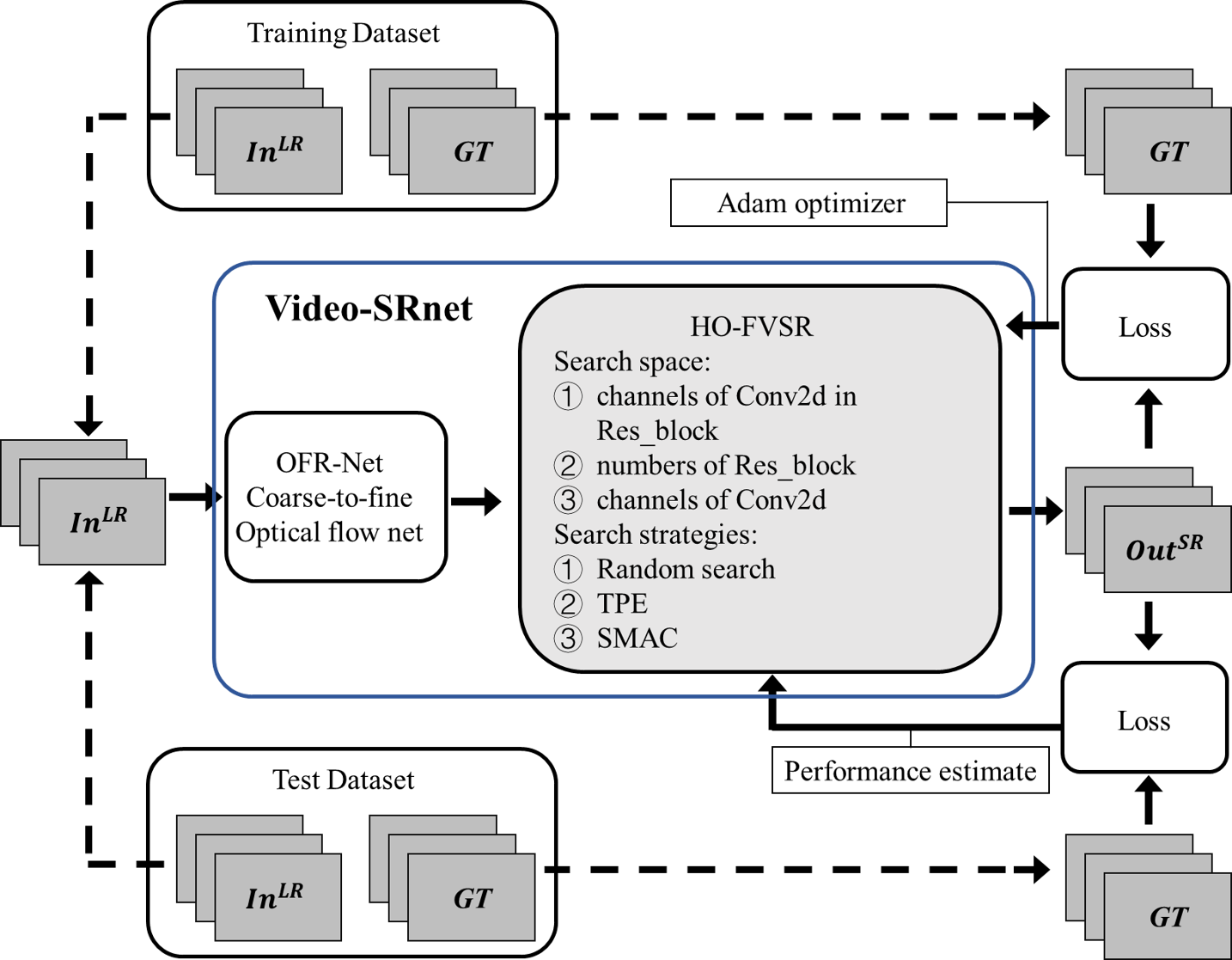}
   \caption{Hyper-parameter optimization design for face video super-resolution network. The training process uses training dataset and adam optimizer to tune parameters in HO-FVSR. Our video super-resolution network contains a fixed architecture called OFR-net, which is the same network design in \cite{7}}.
   \label{fig3}
\end{figure}

From the overall framework in Figure \ref{fig2}, it can be seen that the algorithm only performs VSR processing for face images. We now investigate the realization of reducing parameters and operations in VSR network. We define the hyper-parameter optimization process as the interaction between three sub-modules: search space, search strategy and performance estimation, as shown in Figure \ref{fig1}. For face super-resolution optimization, the architecture of network can be inherited from general video algorithms~\cite{6,7}. Only the hyper-parameters in the network need to be automatically searched. The search space used in this work are defined as the number of channels of the convolution kernel in the residual submodule, the number of residual submodules, and the number of convolution kernel channels in the up-sampling submodule. The search strategy uses three existing methods: TPE~\cite{8}, random search~\cite{9} and SMAC~\cite{10}. In training process, we divide dataset into two parts: training dataset and evaluation dataset, and then use training dataset to conduct preliminary training. We define a fixed epoch number as the end of preliminary training, and then switch to the evaluation dataset for performance evaluation. The search strategy will select the next candidate hyper-parameter in the search space according to the obtained evaluation performance.
%until the parameter optimization algorithm ends. 
The block diagram of hyper-parameter optimization design is shown in Figure \ref{fig3}.

\begin{figure*}[ht]
   \centering
   \includegraphics[width=\textwidth]{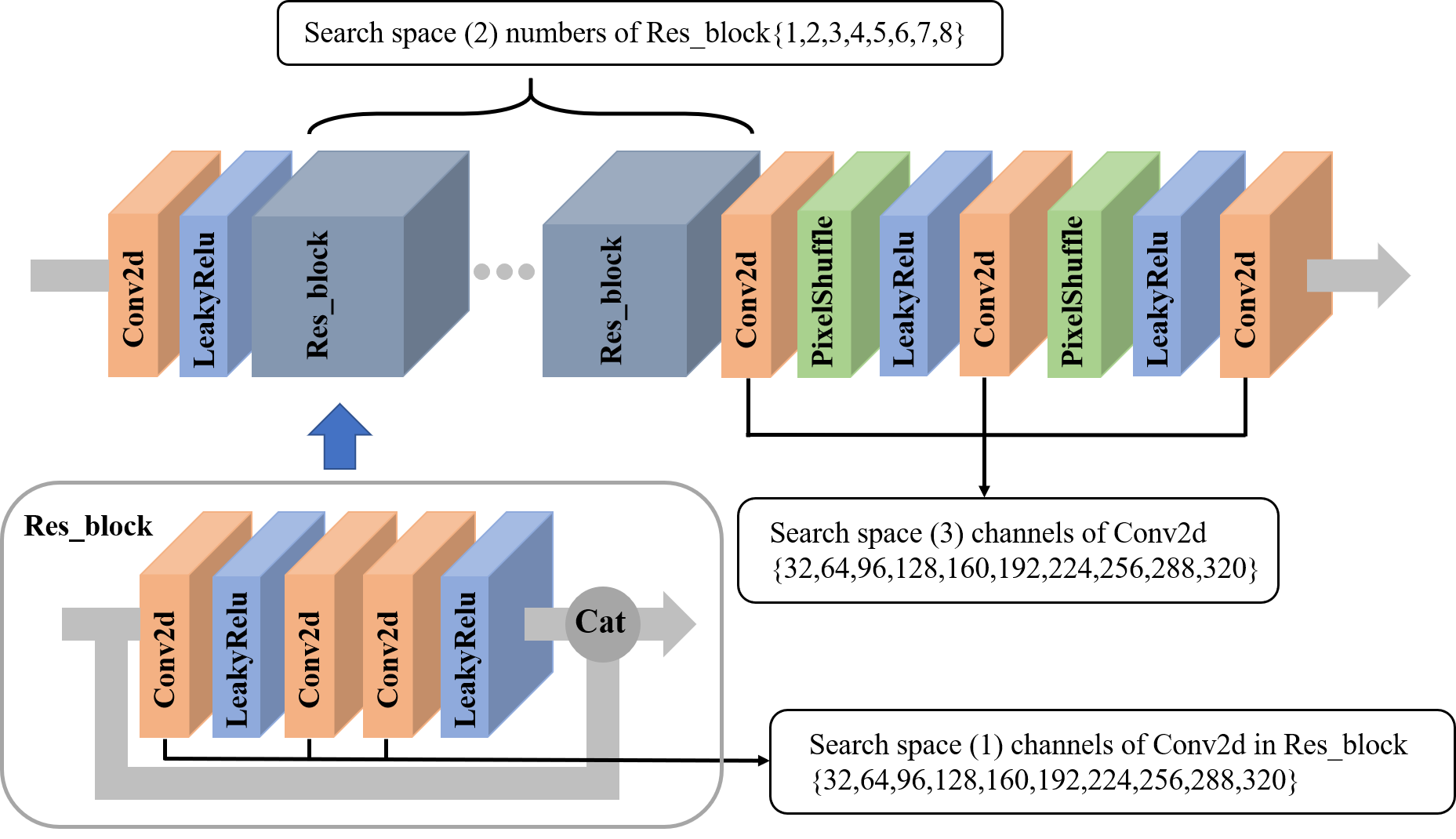}
   \caption{The network architecture of HO-FVSR. The architecture design consists of residual submodules and up-sampling submodules.}
   \label{fig4}
\end{figure*}

The HO-FVSR network architecture is shown in Figure \ref{fig4}. Each search space parameter has multiple candidate values: 1) channels of Conv2d in Res\_block lie in \{32, 64, 96, 128, 160, 192, 224, 256, 288, 320\}; 2) numbers of Res\_block lie in \{1, 2, 3, 4, 5, 6, 7, 8\}; 3) channels of Conv2d lie in \{32, 64, 96, 128, 160, 192, 224, 256, 288, 320\}. There are 800 possible combinations of three candidate values. Iterating through all the possible situations will be time-consuming. In our experiments, we set 40 hyper-parameter sets as search count for each search strategy, and set the running time of each search strategy to not exceed 32 hours. To address the time-consuming process of optimization, we design a simultaneous train-evaluation method for acceleration while ensuring network training convergence. 

The method is depicted in Table \ref{tab1}. We set 20 epochs as training epoch number to make a trade-off between training time and performance estimation. The reason why a smaller epoch number was set is that the more the number of training epochs, the more likely the network will tend to converge. However, the time consumption will be greater, making the optimization process slower.
\begin{table}[h!]
   \caption{Simultaneous train-evaluation method.}
   \label{tab1}
   \begin{tabular}{m{\columnwidth}}
   \toprule
   Hyper-parameter Optimization for Face Video Super-Re- solution (HO-FVSR)                                                         \\
   \midrule
   Define search space $A$, search strategy $S$, hyper-parameter $H$, search count $T_1$, training epoch number $T_2$, model $f$. \\
   HO-FVSR:                                                                                                                       \\
   1. Initialize a hyper-parameter $H$ to model $f: S(A) \rightarrow H$;                                                          \\
   2. For $t_{1} \leftarrow 0$ to $T_{1}$:                                                                                        \\
   3. \quad For $t_{2} \leftarrow 0$ to $T_{2}$:                                                                                        \\
   \qquad a) Train $f(H)$ based on training dataset;                                                                                     \\
   \qquad b) Evaluate $f(H)$ based on evaluation dataset;                                                                                \\
   \qquad c) Record the evaluation loss, $L_{-}\textnormal{eval}$;                                                                                    \\
   4. \quad Select a new hyper-parameter $H^{\prime}$ to model $f$,                \\
   \qquad $S\left(A, L_{-}\textnormal{eval}\right) \rightarrow H^{\prime}$;\\
   5. Choose the best hyper-parameter $H^{\ast}$ based on evaluation              \\ 
   \quad loss. \\
   \bottomrule                                        
   \end{tabular}
\end{table}

\section{Results}
\begin{figure*}[ht]
   \centering 
   \includegraphics[width=\textwidth]{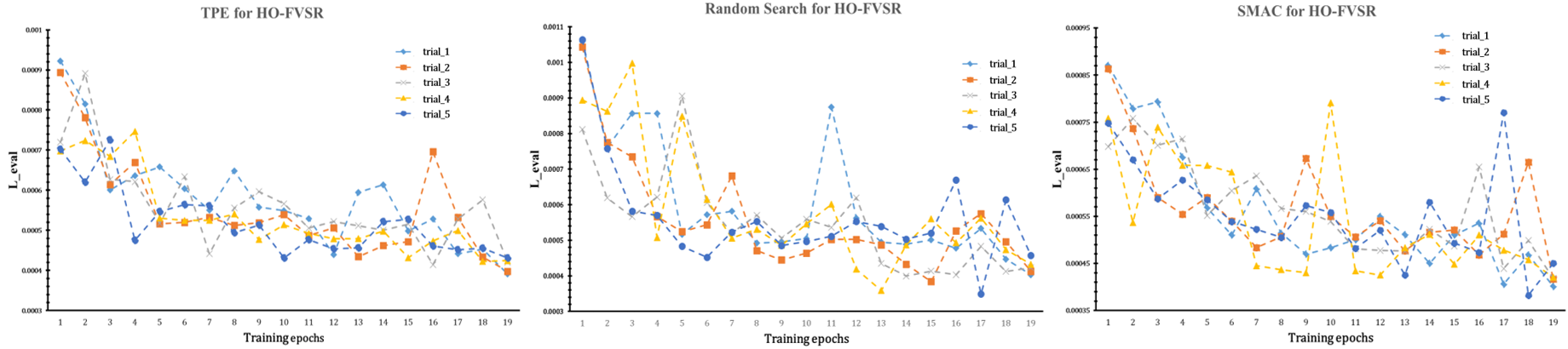}
   \caption{Top 5 networks for each search strategy. The ordinate represents the error of the network on the evaluation dataset.}
   \label{fig5}
\end{figure*}

We conducted experiments on GTX1080Ti, using Pytorch deep learning development framework and NNI auto machine learning tool provided by Microsoft which can record real-time training messages.
For each search strategy, we listed top 5 networks in the performance evaluation, and recorded their performance changes during training process, as shown in Figure \ref{fig5}. It can be seen that as the number of training epochs increases, the error of the neural network in evaluation dataset shows a decreasing trend. Although the performance of the network fluctuates during the training process, the overall trend shows convergence. At the same time, as the training time increases, the performance improvement has slowed down. It can be clearly seen that after the $10^\textnormal{th}$ epoch cycle, the rate of error reduction is gradually decreasing. 

Table \ref{tab2} shows the time consumption required by the simultaneous train-evaluation method described in Table \ref{tab1}. We set each training strategy to list no more than 40 sets of hyper-parameters in search space, the corresponding total training time does not exceed 32 hours. When each network associated with selected hyper-parameters trains for 20 epochs, it will switch to the next set of candidate hyper-parameters. As can be seen from Table \ref{tab2}, three search strategies selected about 23 network hyper-parameters respectively, which took less than 32 hours, and the total time was less than 100 hours. It usually takes about 200 epochs to train a network to fully converge within 18 hours. If all 800 candidate hyper-parameters in the search space are trained to fully convergence, it would elapse about 14,400 hours. In this perspective, simultaneous train-evaluation method has greatly accelerated the process of overall optimization process.
\begin{table}[h!]
   \caption{Time consumption for simultaneous train-evalua- tion method.}
   \label{tab2}
   \begin{tabular}{cccc}
   \toprule
   Strategy              & Networks & Epochs & Time      \\
   \midrule
   TPE~\cite{8}           & 24       & 20     & 31h 31min \\
   Random search~\cite{9}  & 22       & 20     & 29h 54min \\
   SMAC~\cite{10}          & 23       & 20     & 30h 11min \\
   Efficient network     & 1        & 200    & 18h 30min\\ 
   \bottomrule
   \end{tabular}
\end{table}

\begin{figure}[h!]
   \centering 
   \includegraphics[width=0.94\columnwidth]{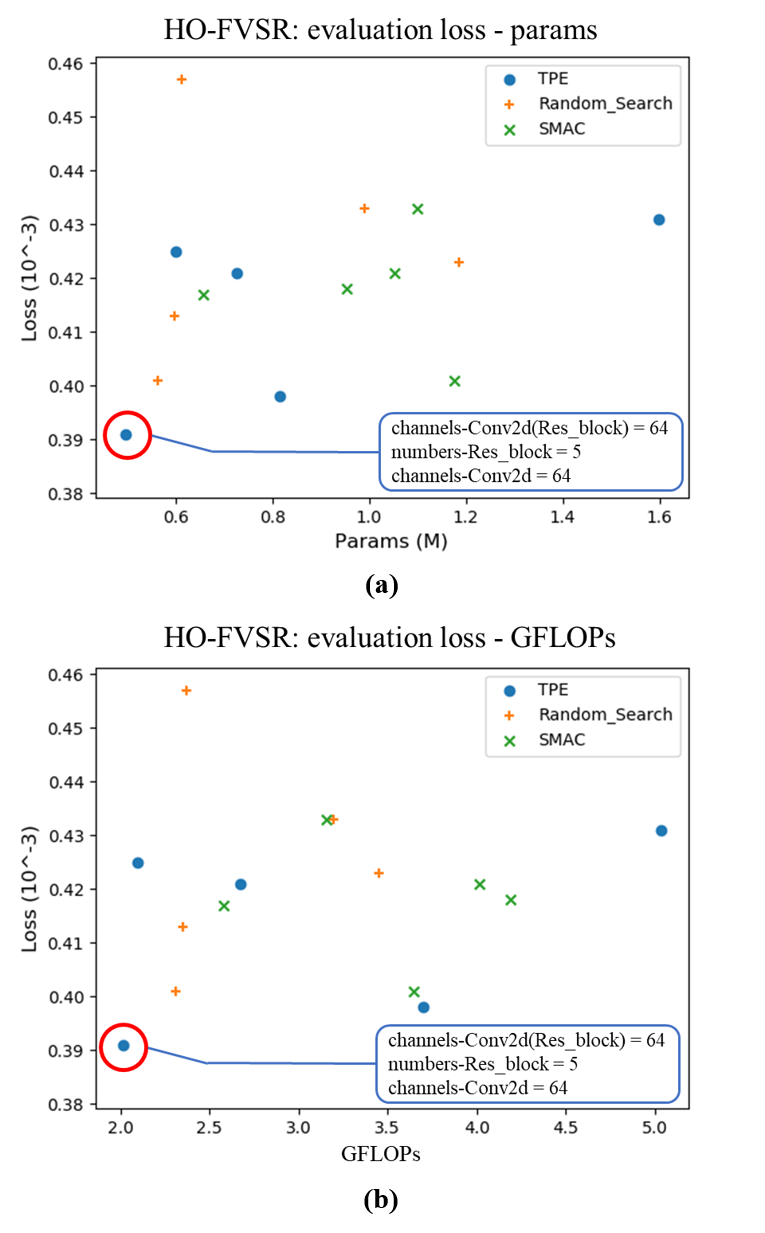}
   \caption{Scatter plot of the parameters, operations and performance of the candidate network. The parameter quantity in figure (a) is based on M (10\^{}6). The GLOFs in figure (b) is the abbreviation of Giga Float-point Operations (10\^{}9).}
   \label{fig6}
\end{figure}

For the top 5 networks in each search strategy, we calculated the number of parameters and operations performed in face video super-resolution, and represented them in a scatter diagram, as is shown in Figure \ref{fig6}. We assume that the resolution of input video sequence is $36\times 36\times 1$. Three consecutive frames of images are needed when performing face video super resolution. Therefore, the operations in Figure \ref{fig6} is calculated according to a vector of input size $36\times 36\times 1\times 3$.

\begin{figure*}[ht]
   \centering 
   \includegraphics[width=\textwidth]{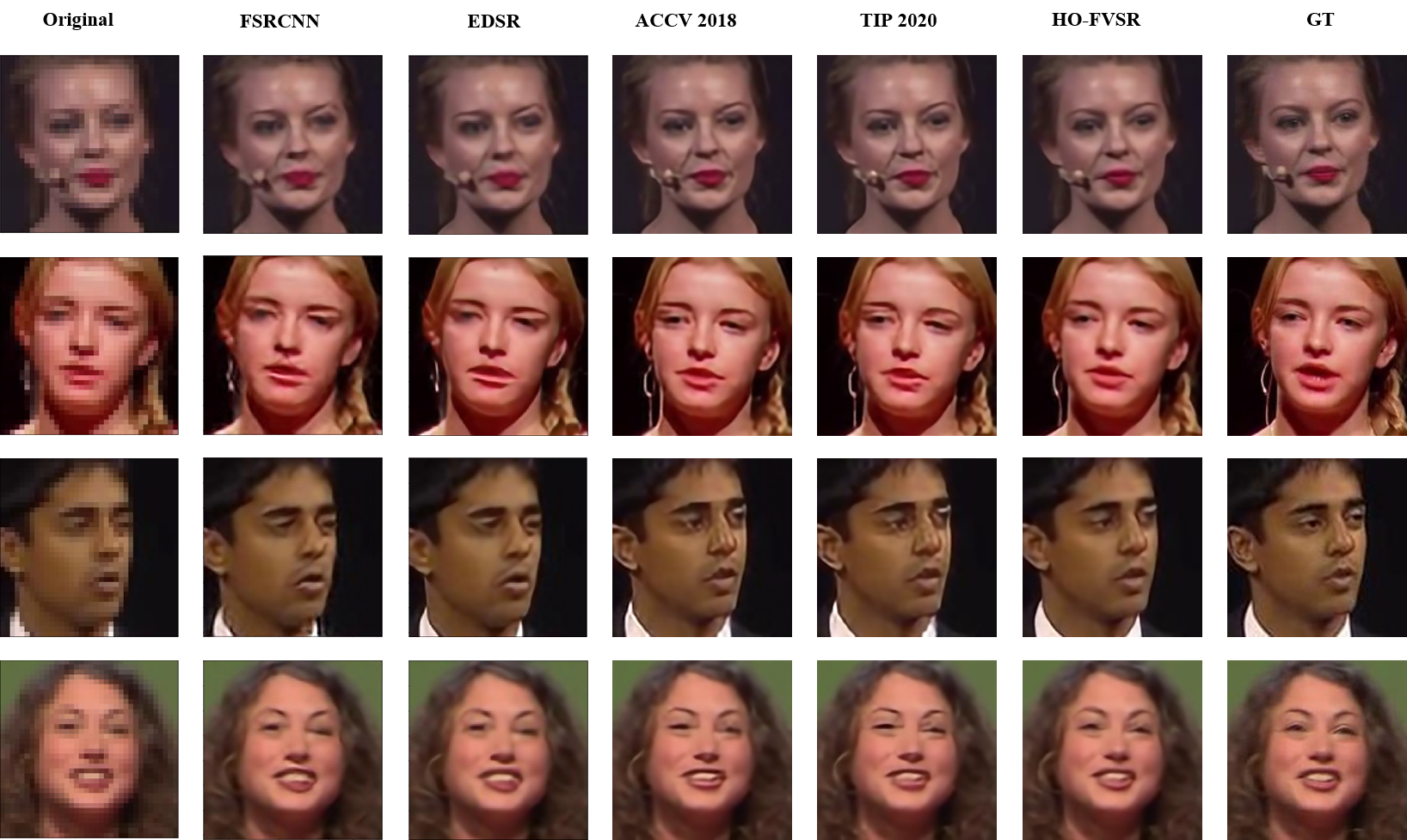}
   \caption{The comparison between FSRCNN~\cite{4}, EDSR~\cite{5}, ACCV-2018~\cite{6}, TIP-2020~\cite{7} and HO-FVSR.}
   \label{fig7}
\end{figure*}

Based on the results in Figure \ref{fig6}, it can be seen that TPE search strategy can provide an efficient hyper-parameter solution with advantages in estimated performance, parameters, and operations: \{channels of Conv2d in Res\_block = 64; numbers of Res\_block = 5; channels of Conv2d = 64\}. This set of hyper-parameters is selected and trained on the training dataset for 200 epochs to obtain the final efficient network for face video super-resolution tasks. We compare the HO-FVSR with other state-of-the-art images and video super-resolution networks, and the results are shown in Figure \ref{fig7} and Table \ref{tab3}.
\begin{table}[h!]
   \centering
   \caption{Performance comparison between video super-resolution algorithms.}
   \label{tab3}
   \begin{tabular}{cccc}
   \toprule
              & ACCV-2018\cite{6} & TIP-2020\cite{7} & HO-FVSR        \\
   \midrule
   Params (M) & 1.64              & 1.05             & \textbf{0.50}  \\
   GFLOPs     & 2.55              & 2.74             & \textbf{2.02}  \\
   Speed (s)  & 0.031             & 0.024            & \textbf{0.021} \\
   FPS        & 32.26             & 41.67            & \textbf{47.62} \\
   PSNR (dB)  & 29.25             & 29.00            & \textbf{30.14} \\
   SSIM       & 0.898             & 0.890            & \textbf{0.909}\\ 
   \bottomrule
   \end{tabular}
\end{table}

Compared with HO-FVSR, super resolution algorithms based on single image may introduce abnormal super-resolution results compared with normal facial image contours, which cannot reflect the true restoration of the original image. This is due to the limited input information in a single image based on super-resolution algorithms. 
The redundant information between continuous video frames cannot be effectively extracted and utilized. Video-based super-resolution algorithms  show better human visual perception characteristics, and the details are restored realistic. In Table \ref{tab3}, it can be seen that our HO-FVSR can achieve at least 52.4\% parameters and 20.7\% FLOPs reduction versus previous works~\cite{6,7}. 
HO-FVSR also has advantages in speed, peak signal-to-noise ratio (PSNR), and structural similarity (SSIM) when performing $36\times 36\times 1\times 3$ video sequence super-resolution, benefiting from the reduction of model parameters and operations. The proposed HO-FVSR improves the PSNR and SSIM by approximately 1dB and 0.01 respectively, reaches 47.62 images processing per second, which meets real-time video processing requirements. However, even if the number of operations and parameters are significantly reduced, the network speed does not increase correspondingly. This may be caused by some execution bottlenecks in network layers.

\section{Conclusion}
In this paper, we created a specific dataset consisting entirely of face video sequences for face super resolution network training and evaluation, meanwhile conduct hyper-parameter optimization in our experiments. We use three combined strategies to optimize the network parameters with a simultaneous train-evaluation method to accelerate optimization process. Results show that simultaneous train-evaluation method helps to make networks converge and generate most efficient network. Our efficient network (HO-FVSR) can reduce at least 52.4\% parameters and 20.7\% FLOPs, achieve promising performance on execution time, PSNR and SSIM compared with state-of-art video super-resolution algorithms. In the future, we will investigate the benefits from reconfigurable hardware architectures for dynamic network architecture designs.

{\small
\bibliographystyle{ieee_fullname}
\bibliography{egbib}
}

\end{document}